\documentclass{article}
\usepackage{spconf,amsmath,graphicx}
\usepackage{multirow}
\usepackage{subcaption}
\usepackage{url} 
\usepackage{todonotes}

\title{Open Implementation and Study of BEST-RQ for Speech Processing}
\name{Ryan Whetten$^1$, Titouan Parcollet$^2$, Marco Dinarelli$^3$, Yannick Estève$^1$ 
\thanks{This work received funding from the French ANR
E-SSL project (grant N°ANR-22-CE23-0013) and was performed using HPC resources from GENCI–IDRIS (projects A0131013821 and AD011014732)}
}
\address{$^1$ Laboratoire Informatique d'Avignon, Avignon Université, France \\
$^2$ Samsung AI Center, Cambridge, United Kingdom \\ 
$^3$ Univervisté Grenoble Alpes, Inria, CNRS, Grenoble INP, LIG, 38000, Grenoble, France}
\begin{document}
%
\maketitle
\begin{abstract}

Self-Supervised Learning (SSL) has proven to be useful in various speech tasks. However, these methods are generally very demanding in terms of data, memory, and computational resources. 
BERT-based Speech pre-Training with Random-projection Quantizer (BEST-RQ), is an SSL method that has shown great performance on Automatic Speech Recognition (ASR) while being simpler than other SSL methods, such as wav2vec 2.0. 
Despite BEST-RQ's great performance, details are lacking in the original paper, such as the amount of GPU/TPU hours used in pre-training, and there is no official easy-to-use open-source implementation. Furthermore, BEST-RQ has not been evaluated on other downstream tasks aside from ASR and speech translation. 
In this work, we describe a re-implementation of a Random-projection quantizer and perform a preliminary study with a comparison to wav2vec 2.0 on four downstream tasks. We discuss the details and differences of our implementation. We show that a random projection quantizer can achieve similar downstream performance as wav2vec 2.0 while decreasing training time by over a factor of two.


\end{abstract}
\section{Introduction}
\label{sec:intro}

Self-supervised learning (SSL) is a training regime in which a model extracts targets from the input data itself and is trained to predict these targets. SSL systems can thus take advantage of large amounts of unlabeled data in a pre-training stage and use a minimal amount of labeled data to achieve very impressive results on a wide variety of tasks~\cite{mohamed2022self}. 
In the speech domain, SSL has lead to state-of-the-art (SOTA) results in tasks such as Automatic Speech Recognition (ASR), automatic emotion recognition (ER), automatic speaker verification (ASV), Spoken Language Understanding (SLU)~\cite{mohamed2022self,yang2021superb}.

However, SSL pre-training is very data, memory, and computationally expensive. For example, authors of wav2vec 2.0, reported using around 2,400 V100 GPU hours and a batch size of 1.6 hours just for the base model~\cite{baevski2020wav2vec}. On the high end for very large datasets such as \emph{LeBenchmark}, authors have reported using 54,600 A100 GPU hours for their extra-large model~\cite{parcollet2023lebenchmark}. 

Despite efforts to enhance the efficiency of other widely used SSL models for speech, such as HuBERT and data2vec ~\cite{chen2023reducing, baevski2023efficient}, the process remains resource intensive. For instance for HuBERT and \emph{data2vec} it requires around 1000 and 700 A100 GPU hours respectively.

One reason of this high cost is due to the Acoustic Feature Extractors (AFE), which is commonly implemented as a series of Convolutional Neural Network (CNN) layers in SOTA architectures, but recent studies have shown that these can be replaced with more efficient alternatives while maintaining performance~\cite{parcollet2023efficiency}. 

A recent model, BERT-based Speech pre-Training with Random-projection Quantizer (BEST-RQ), reduces this cost simply by using Mel filterbanks in place of the AFE, still obtaining impressive results on ASR~\cite{chiu2022self}. Due to this fact, and a few other simplifications (see Section~\ref{ssec:w2v2-vs-bestrq}), BEST-RQ appears to be one of the most efficient SSL methods proposed until now.
Currently, no official implementation of BEST-RQ exists, inhibiting the community from having access to a very efficient SSL training method. Furthermore, BEST-RQ performance has only been investigated on two tasks, ASR and speech translation~\cite{zhang2023google}. We aim to fill in these gaps.

In this work, we present our open-source implementation of a random projection quantizer using SpeechBrain~\cite{ravanelli2021speechbrain} and conduct an initial examination by comparing BestRQ to wav2vec 2.0.\footnote{The code for our BEST-RQ implementation is available at \url{https://github.com/speechbrain/speechbrain/tree/develop/recipes/LibriSpeech}} 
We analyze the pre-training time as well the performance on the following downstream tasks: automatic speech recognition (ASR), automatic speaker verification (ASV), intent classification (IC), and emotion recognition (ER).  
Results show that a random projection quantizer can achieve similar performance on these various downstream tasks as wav2vec 2.0 with the added benefit of reducing SSL pre-training time by more than half. 
We believe our open-source implementation can serve as a starting point for furthering research, facilitating exploration into various SSL architectures for speech.

\section{Background}
\label{sec:background}
\vspace{-0.3cm}

While there are other effective SSL models for speech processing, such as HuBERT and data2vec ~\cite{chen2023reducing, baevski2023efficient}, wav2vec 2.0 seems more wide-spread and used by the speech community, with several freely available tools and models, making thus a comparison straightforward.
In this section, we give a more detailed comparison of wav2vec 2.0 and BEST-RQ.

\vspace{-0.3cm}

\subsection{wav2vec 2.0 vs. BEST-RQ}
\label{ssec:w2v2-vs-bestrq}


\begin{figure*}
  \begin{subfigure}{0.5\textwidth}
    \centering
    \includegraphics[width=\linewidth]{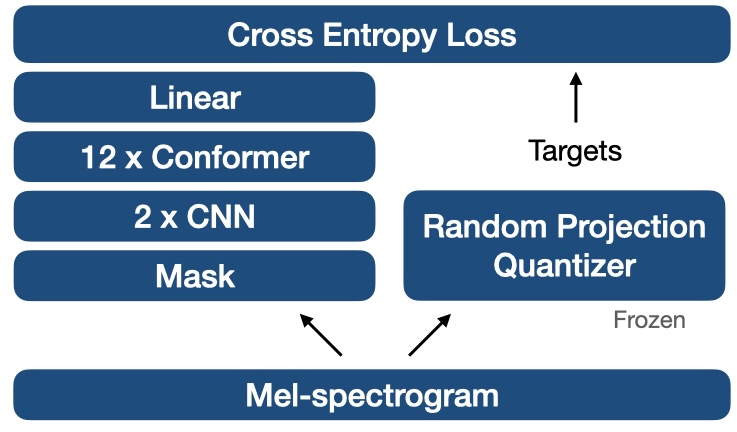}
    \caption{BEST-RQ}
    \label{fig:brq-arch}
  \end{subfigure}%
  \begin{subfigure}{0.5\textwidth}
    \centering
    \includegraphics[width=\linewidth]{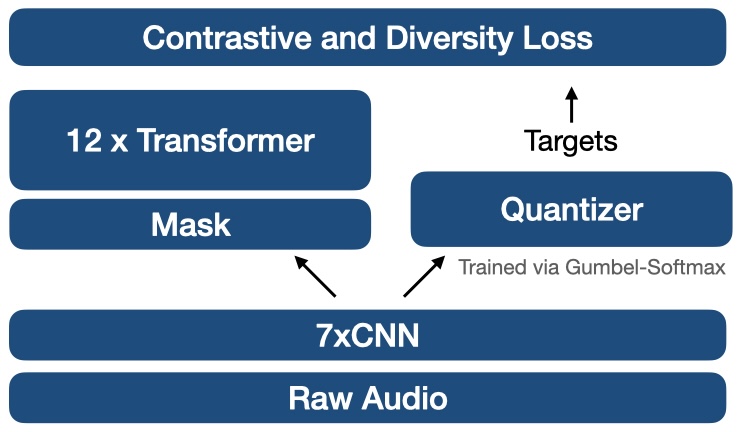}
    \caption{wav2vec 2.0}
    \label{fig:w2v2-arch}
  \end{subfigure}
  \caption{Diagrams of our BEST-RQ and wav2vec 2.0 architecture. BEST-RQ operates on mel-spectrograms, uses a static quantizer and conformer layers. On the other hand, wav2vec 2.0 operates on raw audio, trains the quantizer, and uses transformer layers.}
  \label{fig:arch}
\end{figure*}

BEST-RQ and wav2vec 2.0 have a few major differences, one of these being that \textbf{BEST-RQ does not operate on the raw audio}. Instead, BEST-RQ operates on 80-dimensional log Mel filterbank coefficients with two CNN layers for the Acoustic Feature Extractor (AFE). In contrast, wav2vec 2.0 operates on the raw audio using seven CNN layers for the AFE.

Using a hand-crafted AFE such as Mel filterbanks significantly reduces the amount of memory and computations needed to train the model. However, this may limit the performance of the model as filterbanks are compressed views of speech based on psycho-acoustics (i.e. based on human expertise) and, consequentially, could remove acoustic cues that a non-hand-crafted AFE can learn. 

A second major difference is that \textbf{BEST-RQ uses a randomly initialized linear layer and one codebook for quantizing and discretizing the audio}. These components are both fixed throughout training. The Mel filterbanks are projected with the linear layer, then the index of the nearest codebook entry to the projection is used as the target. The nearest codebook entry is found by taking the argmin of the normalized distance between each entry in the codebook and the projection. 

Afterward a mask is applied to a portion of Mel filterbanks and the objective of the model is to guess the correct targets for the masked sections. This is treated as a classification task using cross entropy to calculate the loss. 

Conversely, in wav2vec 2.0, there are two learned codebooks which are trained via a gumbel softmax. The loss function for wav2vec 2.0 is more complicated consisting of a combination of a contrastive loss (i.e. identifying the correct label among a portion of randomly selected false labels) and a diversity loss (to prevent the model from collapsing to using just one entry in the codebook). 

One last major difference is that \textbf{BEST-RQ uses conformer layers} instead of transformer layers. Note that although conformer layers with wav2vec 2.0 has been studied \cite{zhang2020pushing}, for this work we only study the wav2vec 2.0 standard architecture as our goal is not to prove one methodology is better than another, it is instead to show that our implementation works as expected and compares favorably to a well-known alternative. See Figure~\ref{fig:arch} for a simple diagram comparing these two architectures.

\cite{chiu2022self} showed that good ASR performance (in terms of word error rates) can be achieved without having to train an AFE or codebooks. Theoretically, this significantly reduces the complexity of the model, the loss function and the backward pass. As the pre-training time was not reported in the original BEST-RQ paper, we aim to show how this affects pre-training empirically and offer evaluations on ASR and three other downstream tasks. For reproducibility and to allow others to study effects of using a random projection quantizer, we make our implementation open source.

\section{Experiments}
\label{sec:methods}
\vspace{-0.3cm}
For our experiments we pre-train a wav2vec 2.0 model and our BEST-RQ implementation and then evaluate on a subset of the MP3S benchmark tasks~\cite{zaiem2023speech}. In the next sections we describe the pre-training environment, the architecture of the models, and the downstream tasks.

\subsection{Pre-training Setting}
\label{ssec:pre-train}
For pre-training, we use 960 hours of LibriSpeech (i.e. the \emph{train-clean-100}, \emph{train-clean-360} and \emph{train-other-500} splits)~\cite{panayotov2015librispeech}. 
We train all models for 42 epochs or roughly 200k steps using eight Tesla V100 GPUs. We save a checkpoint at epoch 21, around 100k steps, to evaluate performance throughout training. We denote the difference between these versions of the model by naming them \emph{100k} and \emph{200k}. 

We use the same dynamic batching settings for both wav2vec 2.0 and BEST-RQ with the max batch size being 100 seconds. Because we use eight GPUs, the total batch size is 800 sec or about 13.33 minutes. We chose this batch size to be able to run our experiments on small GPUs and within a limited amount of time using only a few hundred GPU hours.


\subsection{Model Settings}
\label{ssec:models}
We use the wav2vec 2.0 base model architecture~\cite{baevski2020wav2vec} as a baseline, using the implementation of SpeechBrain. We only modified the dynamic batch settings to have a max batch size of 100 seconds.

For our BEST-RQ implementation, we follow overall the author's description, but we make a \textbf{few key changes} found through some preliminary experiments, and that revealed very important for performance with the relatively small batch size (13 min. vs. 18hours), small amount of pre-training data (960 hours vs. for example 12 million \cite{zhang2023google}), and number of GPUs. We \textbf{shrink the number of conformer layers from 24 to 12}. This was done in order to match the same number of layers as wav2vec 2.0 base model, and allowed the model to fit on lower-end GPUs. We \textbf{added layer dropout of 0.05}, we \textbf{reduced the learning rate to 0.0008} and \textbf{lastly we increase the mask ratio to about 60\% of the audio} (i.e. 15\% of frames are randomly selected to be masked together with the following three frames). We describe some results from our initial experiments in Section~\ref{ssec:prelim}. 

\subsection{Downstream Tasks}
\label{ssec:downstream}
For the downstream tasks we follow the methodology in the MP3S~\cite{zaiem2023speech} and the SUPERB~\cite{yang2021superb} benchmarks, that is the SSL model is frozen and the input to the downstream task specific model is a weighted sum of the hidden states from the pre-trained model. In the MP3S, each downstream task has two probes, of which we select one per task due to time and resource constraints. 


For \textbf{Automatic Speech Recognition (ASR)} we use the \emph{train-clean-100/dev-clean} for training and validation respectively and then evaluate on the \emph{test-clean} and \emph{test-other splits} from LibriSpeech~\cite{panayotov2015librispeech}. 

For this task, we use a 2-layered BiLSTM followed by a linear layer, softmax layer, and then CTC loss~\cite{zaiem2023speech}. The BiLSTM has a size of 1024 and a dropout of 0.2. The metric we use to measure performance is the Word Error Rate (WER). We report the WER with and without applying the official 4-gram ARPA LM~\cite{zaiem2023speech}.\footnote{Available at \url{openslr.org/11/}}

For \textbf{Automatic Speaker Verification (ASV)} we use Voxceleb1 \cite{nagrani2017voxceleb}. This dataset is made up of utterances of over 1,000 celebrities gathered from youtube. The task is a binary classification task where given two audio files, the model has to determine whether the speaker is the same or not. 

We use the ECAPA-TDNN probe from the MP3S benchmark. An ECAPA-TDNN uses a time delayed neural network, statistical pooling with temporal and channel self-attention in order for generating fixed length high quality speaker embeddings~\cite{desplanques2020ecapa}. To measure the performance of ASV we use the equal error rate (EER).

For \textbf{Intent Classification (IC)} we use SLURP~\cite{bastianelli-etal-2020-slurp}, which is known to be more challenging than other datasets, consisting of 177 speakers from 72k audio files adding up to 58 hours of audio. The tasks is to classify a given utterance into 1 of 18 tasks or scenarios such as \emph{email}, \emph{calendar}, or \emph{play} (as in \emph{play next song}). 

For the probe, again we use one from MP3S. This probe is a BiLSTM with a size of 1024 that is followed by a linear layer, a statistic pooling layer, and then a linear layer for the final classification. The evaluation metric for this task is the accuracy.

For \textbf{Emotion Recognition (ER)} we use the IEMOCAP dataset \cite{busso2008iemocap}. This dataset contains about 12 hours of data from 10 speaker actors performing scripts with four different emotions (neutral, happy, sad and angry). The performance is measured with a 10-fold cross validation. Similar to the ASV task we use the ECAPA-TDNN probe from the MP3S benchmark.



In order to explore the capabilities of BEST-RQ not as just a model for speech embedding but as a model that can be fine-tuned (i.e. not frozen) for a downstream task, we perform one more experiment where we fine-tune the model using a vanilla feed-forward neural network with CTC loss.
We fine-tune using the \emph{train-clean-100} and then evaluate on the \emph{test-clean} and \emph{test-other} splits from LibriSpeech with and without the 4-gram ARPA language model. 



\begin{table*}[t]
    \centering
    \begin{tabular}{cccccccccc}
        \hline
        \textbf{Model / Task}
        & \textbf{\# Par.}
        & \textbf{GPU hours}
        & \multicolumn{4}{c}{\textbf{LibriSpeech train-100 ASR}} 
        & \textbf{VoxCeleb1}
        & \textbf{SLURP} 
        & \textbf{IEMOCAP} \\
        Metric &&& \multicolumn{4}{c}{WER $\downarrow$} 
        & EER $\downarrow$
        & Acc. $\uparrow$
        & Acc. $\uparrow$\\
        &&&                      Clean & Clean LM & Other & Other LM &  ASV &  IC  & ER   \\
        \hline
        W2V2 100k & 90.9M & 130 & 16.26 &  10.63  & 40.17 &  30.83   & 4.56 & 72.8 & 60.9 \\
        W2V2 200k & 90.9M & 262 & 13.89 & 9.45     & 33.55 & 25.49    & 3.83 & 74.5 & 63.0 \\
        BRQ 100k  & 83.0M &  54 & 16.79 & 10.79     & 38.09 & 28.31    & 3.84 & 74.3 & 61.3 \\
        BRQ 200k  & 83.0M & 109 & 15.11 & 9.76     & 34.06 & 24.74    & 3.53 & 74.8 & 63.8 \\

        \hline
    \end{tabular}
    \caption{Results on downstream tasks at 100k and 200k steps. BEST-RQ performs similarly to wav2vec but with less than half of the training time. Training time and number of parameters are shown under \emph{GPU hours} and \emph{\# Par.} respectively.}
    \label{tab:mp3-results}
\end{table*}


\section{Results}
In this section we report results on the pre-training time, downstream tasks, and on some initial experiments varying the masking ratio and the size of the codebook in our BEST-RQ implementation.
\label{sec:results}


\subsection{MP3S Downstream Tasks and Fine-Tuning}

Among downstream experiments, reported in Table~\ref{tab:mp3-results}, wav2vec 2.0 proved to perform better than BEST-RQ without a language model on the MP3S LibriSpeech \emph{train-100} ASR task. However, with the language model, BEST-RQ performs very close to wav2vec 2.0 on the \emph{test-clean} and even slightly better on the \emph{test-other} split. Additionally, when we fine-tune the model for ASR, BEST-RQ performs better than wav2vec 2.0. On the other tasks, ASV, IC and ER, BEST-RQ performs slightly better than wav2vec 2.0, while pre-training approximately 2.4 times faster than wav2vec 2.0.

Examining the difference between the models at 100k and 200k training steps, in Table~\ref{tab:ft-results}, we note that in every case wav2vec 2.0 has a greater improvement in performance. For example on IC, the wav2vec 2.0 200k model improves by 1.7\% over the 100k model, while BEST-RQ improves by 0.5\%. We believe this is due to the fact that the AFE in wav2vec 2.0 has to be trained, which results in slower convergence in terms of both time and number of steps.

\begin{table}[h]
    \centering
    \begin{tabular}{ccccccc}
        \hline
        \textbf{Model}
        & \multicolumn{4}{c}{\textbf{Fine-Tune LibriSpeech train-100}}  \\
        &          Clean & Clean LM & Other & Other LM \\
        \hline
        W2V2 100k & 16.42 & 10.29 & 36.62 & 27.25    \\
        W2V2 200k & 13.47 &  8.73 & 29.64 & 21.92    \\
        BRQ 100k  & 14.59 &  9.00 & 31.49 & 23.04    \\
        BRQ 200k  & 12.21 &  7.78 & 26.81 & 19.67    \\
        \hline
    \end{tabular}
    \caption{Results of fine-tuning on ASR with LibriSpeech \emph{train-100}.}
    \label{tab:ft-results}
\end{table}
\vspace{-0.5cm}

\subsection{Impact of the Masking Ratio}
\label{ssec:prelim}

 We train all models for 18 epochs or about 87k steps on 4 x 11GB 2080Ti GPUs and we only looked at the ASR downstream performance on the validation set (i.e. the \emph{dev-clean} split of LibriSpeech). We combined the variation of the masking ratio with the variation of the codebook size when using 1\% and 10\% masking ratios.

As shown in Table~\ref{tab:mask-cb}, shrinking the codebook (CB) from the original 8192 to 1024 does not seem to change the downstream performance in any consistent or significant way. At a masking ratio of 1\% the smaller codebook performs slightly better, while at 10\% the smaller codebook performs slightly worse. Instead, increasing the amount of masked frames has a great impact, decreasing the WER on the \emph{dev-clean} split by about 15\% (from a WER around 35\% down to a bit above 20\%) as the amount of frames selected for masking increased from 1\% to 12\%. 

\begin{table}[h]
    \centering
    \begin{tabular}{ccc}
        \hline
        \textbf{Mask \%} & \textbf{Dev-Clean WER} & \textbf{CB} \\
        \hline
        ~1\% & 34.08 & 1024 \\
        ~1\% & 36.45 & 8192 \\
        ~5\% & 25.24 & 8192 \\
        10\% & 21.10 & 1024 \\
        10\% & 20.68 & 8192 \\
        12\% & 20.11 & 8192 \\
        \hline
    \end{tabular}
    \caption{Impact of changing the masking ratio and the codebook size. The mask \% concerns the start frame index chosen for the mask. Since the following three frames are also masked the actual masking ratio is 4 times as big.}
    \label{tab:mask-cb}
\vspace{-0.5cm}
\end{table}
\vspace{-0.3cm}

\section{Discussion and Conclusion}
\vspace{-0.3cm}

\label{sec:conclusion}
In this work we described our open-source implementation of a random projection quantizer and we compared it to wav2vec 2.0 in terms of pre-training time and performance on various downstream tasks. BEST-RQ demonstrates comparable performance to wav2vec 2.0 while speeding up pre-training time by more than half. We will release our code in the \emph{SpeechBrain} toolkit~\cite{ravanelli2021speechbrain}, and believe our open-source implementation can allow researchers to further investigate these methodologies and explore diverse facets of SSL models for speech. 

We hypothesize that because BEST-RQ starts with mel-spectrograms and is about 8M parameters smaller, BEST-RQ converges much faster than wav2vec 2.0 resulting in comparable performance in 200k. The differences in performance between the 100k and 200k models suggest that wav2vec 2.0 model could surpass BEST-RQ given more time or other training settings (such as a larger batch size).

Nevertheless, we believe our results to be insightful on the capabilities of these methods given only a few hundred GPU hours and leave more in depth experiments for future work. 

\vspace{-0.3cm}
\section{Acknowledgment}
\vspace{-0.3cm}
We would like to thank the authors of BEST-RQ~\cite{chiu2022self} for their helpful discussion.



\bibliographystyle{IEEEbib}
\bibliography{strings,refs}

\begin{thebibliography}{10}

\bibitem{mohamed2022self}
Abdelrahman Mohamed, Hung-yi Lee, Lasse Borgholt, Jakob~D Havtorn, Joakim Edin, Christian Igel, Katrin Kirchhoff, Shang-Wen Li, Karen Livescu, Lars Maal{\o}e, et~al.,
\newblock ``Self-supervised speech representation learning: A review,''
\newblock {\em IEEE Journal of Selected Topics in Signal Processing}, 2022.

\bibitem{yang2021superb}
Shu-wen Yang, Po-Han Chi, Yung-Sung Chuang, Cheng-I~Jeff Lai, Kushal Lakhotia, Yist~Y Lin, Andy~T Liu, Jiatong Shi, Xuankai Chang, Guan-Ting Lin, et~al.,
\newblock ``Superb: Speech processing universal performance benchmark,''
\newblock {\em arXiv preprint arXiv:2105.01051}, 2021.

\bibitem{baevski2020wav2vec}
Alexei Baevski, Yuhao Zhou, Abdelrahman Mohamed, and Michael Auli,
\newblock ``wav2vec 2.0: A framework for self-supervised learning of speech representations,''
\newblock {\em Advances in neural information processing systems}, vol. 33, pp. 12449--12460, 2020.

\bibitem{parcollet2023lebenchmark}
Titouan Parcollet, Ha~Nguyen, Solene Evain, Marcely~Zanon Boito, Adrien Pupier, Salima Mdhaffar, Hang Le, Sina Alisamir, Natalia Tomashenko, Marco Dinarelli, et~al.,
\newblock ``Lebenchmark 2.0: a standardized, replicable and enhanced framework for self-supervised representations of french speech,''
\newblock {\em arXiv preprint arXiv:2309.05472}, 2023.

\bibitem{chen2023reducing}
William Chen, Xuankai Chang, Yifan Peng, Zhaoheng Ni, Soumi Maiti, and Shinji Watanabe,
\newblock ``Reducing barriers to self-supervised learning: Hubert pre-training with academic compute,''
\newblock {\em arXiv preprint arXiv:2306.06672}, 2023.

\bibitem{baevski2023efficient}
Alexei Baevski, Arun Babu, Wei-Ning Hsu, and Michael Auli,
\newblock ``Efficient self-supervised learning with contextualized target representations for vision, speech and language,''
\newblock in {\em International Conference on Machine Learning}. PMLR, 2023, pp. 1416--1429.

\bibitem{parcollet2023efficiency}
Titouan Parcollet, Shucong Zhang, Rogier van Dalen, Alberto Gil~CP Ramos, and Sourav Bhattacharya,
\newblock ``On the (in) efficiency of acoustic feature extractors for self-supervised speech representation learning,''
\newblock in {\em Interspeech 2023}, 2023.

\bibitem{chiu2022self}
Chung-Cheng Chiu, James Qin, Yu~Zhang, Jiahui Yu, and Yonghui Wu,
\newblock ``Self-supervised learning with random-projection quantizer for speech recognition,''
\newblock in {\em International Conference on Machine Learning}. PMLR, 2022, pp. 3915--3924.

\bibitem{zhang2023google}
Yu~Zhang, Wei Han, James Qin, Yongqiang Wang, Ankur Bapna, Zhehuai Chen, Nanxin Chen, Bo~Li, Vera Axelrod, Gary Wang, et~al.,
\newblock ``Google usm: Scaling automatic speech recognition beyond 100 languages,''
\newblock {\em arXiv preprint arXiv:2303.01037}, 2023.

\bibitem{ravanelli2021speechbrain}
Mirco Ravanelli, Titouan Parcollet, Peter Plantinga, Aku Rouhe, Samuele Cornell, Loren Lugosch, Cem Subakan, Nauman Dawalatabad, Abdelwahab Heba, Jianyuan Zhong, et~al.,
\newblock ``Speechbrain: A general-purpose speech toolkit,''
\newblock {\em arXiv preprint arXiv:2106.04624}, 2021.

\bibitem{zhang2020pushing}
Yu~Zhang, James Qin, Daniel~S Park, Wei Han, Chung-Cheng Chiu, Ruoming Pang, Quoc~V Le, and Yonghui Wu,
\newblock ``Pushing the limits of semi-supervised learning for automatic speech recognition,''
\newblock {\em arXiv preprint arXiv:2010.10504}, 2020.

\bibitem{zaiem2023speech}
Salah Zaiem, Youcef Kemiche, Titouan Parcollet, Slim Essid, and Mirco Ravanelli,
\newblock ``Speech self-supervised representation benchmarking: Are we doing it right?,''
\newblock {\em arXiv preprint arXiv:2306.00452}, 2023.

\bibitem{panayotov2015librispeech}
Vassil Panayotov, Guoguo Chen, Daniel Povey, and Sanjeev Khudanpur,
\newblock ``Librispeech: an asr corpus based on public domain audio books,''
\newblock in {\em 2015 IEEE international conference on acoustics, speech and signal processing (ICASSP)}. IEEE, 2015, pp. 5206--5210.

\bibitem{nagrani2017voxceleb}
Arsha Nagrani, Joon~Son Chung, and Andrew Zisserman,
\newblock ``Voxceleb: a large-scale speaker identification dataset,''
\newblock {\em arXiv preprint arXiv:1706.08612}, 2017.

\bibitem{desplanques2020ecapa}
Brecht Desplanques, Jenthe Thienpondt, and Kris Demuynck,
\newblock ``Ecapa-tdnn: Emphasized channel attention, propagation and aggregation in tdnn based speaker verification,''
\newblock {\em arXiv preprint arXiv:2005.07143}, 2020.

\bibitem{bastianelli-etal-2020-slurp}
Emanuele Bastianelli, Andrea Vanzo, Pawel Swietojanski, and Verena Rieser,
\newblock ``{SLURP}: A spoken language understanding resource package,''
\newblock in {\em Proceedings of the 2020 Conference on Empirical Methods in Natural Language Processing (EMNLP)}, Bonnie Webber, Trevor Cohn, Yulan He, and Yang Liu, Eds., Online, Nov. 2020, pp. 7252--7262, Association for Computational Linguistics.

\bibitem{busso2008iemocap}
Carlos Busso, Murtaza Bulut, Chi-Chun Lee, Abe Kazemzadeh, Emily Mower, Samuel Kim, Jeannette~N Chang, Sungbok Lee, and Shrikanth~S Narayanan,
\newblock ``Iemocap: Interactive emotional dyadic motion capture database,''
\newblock {\em Language resources and evaluation}, vol. 42, pp. 335--359, 2008.

\end{thebibliography}

\end{document}